\documentclass[sigconf]{acmart}
\usepackage{caption}
\usepackage{subcaption}
\usepackage{booktabs}
\usepackage{multicol}
\usepackage{graphicx}
\usepackage{multirow}
\usepackage{microtype}
\usepackage{makecell}
\AtBeginDocument{%
  \providecommand\BibTeX{{%
    \normalfont B\kern-0.5em{\scshape i\kern-0.25em b}\kern-0.8em\TeX}}}

\copyrightyear{2023}
\acmYear{2023}
\setcopyright{rightsretained}
\acmConference[NumEval-2024]{NumEval-2024}{Draft}{SemEval-2024}
\acmBooktitle{Draft of NumEval-2024}
\acmPrice{15.00}
\acmISBN{978-1-4503-XXXX-XXXXX}

\renewcommand\footnotetextcopyrightpermission[1]{} % removes footnote with conference information in first column

\begin{document}

%\fancyhead{} 
\title{NumHG: A Dataset for Number-Focused Headline Generation}


\author{Jian-Tao Huang}
\affiliation{%
    \institution{Department of Computer Science and Information Engineering, National Taiwan University}
    \city{Taipei}
    \country{Taiwan}}
\email{wzhjttw@gmail.com}

\author{Chung-Chi Chen}
\affiliation{%
  \institution{Artificial Intelligence Research Center, AIST}
  \city{Tokyo}
  \country{Japan}
}
\email{c.c.chen@acm.org}

\author{Hen-Hsen Huang}
\affiliation{%
  \institution{Institute of Information Science, Academia Sinica}
  \city{Taipei}
  \country{Taiwan}
}
\email{hhhuang@iis.sinica.edu.tw}

\author{Hsin-Hsi Chen}
\affiliation{%
    \institution{Department of Computer Science and Information Engineering, National Taiwan University}
    \city{Taipei}
    \country{Taiwan}}
\email{hhchen@ntu.edu.tw}


\begin{abstract}
Headline generation, a key task in abstractive summarization, strives to condense a full-length article into a succinct, single line of text. Notably, while contemporary encoder-decoder models excel based on the ROUGE metric, they often falter when it comes to the precise generation of numerals in headlines. We identify the lack of datasets providing fine-grained annotations for accurate numeral generation as a major roadblock. To address this, we introduce a new dataset, the NumHG, and provide over 27,000 annotated numeral-rich news articles for detailed investigation. Further, we evaluate five well-performing models from previous headline generation tasks using human evaluation in terms of numerical accuracy, reasonableness, and readability. Our study reveals a need for improvement in numerical accuracy, demonstrating the potential of the NumHG dataset to drive progress in number-focused headline generation and stimulate further discussions in numeral-focused text generation.
\end{abstract}

\maketitle
\section{Introduction}
The pursuit of headline generation is an endeavor to distill the essential elements of an article into a single line of text. Though related, this task poses a more significant challenge than merely extracting sentences for summarization, as it requires the crafting of a new sentence encapsulating the same core ideas. As Matsumaru et al.\cite{matsumaru-etal-2020-improving} have demonstrated, the performance of state-of-the-art encoder-decoder models, as judged by the ROUGE metric, is commendable. However, these models sometimes falter by creating inappropriate headlines. The crux of the issue lies in the selection of words that, although superficially similar to the source text, may misrepresent the meaning and be unconnected to the original article. A critical observation from our research is that inaccuracies in the use of "numerals" are a pivotal factor contributing to these erroneous headlines.

\begin{table}[t]
\caption{ An annotation example in NumHG. }
\centering
\resizebox{\columnwidth}{!}{
\begin{tabular}{|p{\columnwidth} |}
\hline 
\textbf{News:} \\
At least \textbf{30} gunmen burst into a drug rehabilitation center in a Mexican border state capital and opened fire, killing \textbf{19} men and wounding \textbf{four} people, police said.
Gunmen also killed \textbf{16} people in another drug-plagued northern city.
The killings in Chihuahua city and in Ciudad Madero marked one of the bloodiest weeks ever in Mexico and came just weeks after authorities discovered \textbf{55} bodies in an abandoned silver mine, presumably victims of the country's drug violence.
More than \textbf{60} people have died in mass shootings at rehab clinics in a little less than \textbf{two} years. Police have said \textbf{two} of Mexico's \textbf{six} major drug cartels are exploiting the centers to recruit hit men and drug smugglers, ...
    \\ \hline
    \textbf{Headline (Question):} 
    Mexico Gunmen Kill \_\_\_\_
    \\
    \hline
    \textbf{Answer}: 35
    \\ 
    \hline
    \textbf{Annotation:} Add(19,16) \\ \hline
\end{tabular}
}
\label{tab:an example in NumHG}
\end{table}

Despite this, datasets that offer fine-grained annotations and frameworks for accurate numeral generation in news headlines are in short supply. In response to this deficit, we propose a novel dataset designed to explore this issue comprehensively. Table~\ref{tab:an example in NumHG} demonstrates an example from our proposed dataset. Our objective is to ensure accurate numeral generation in headlines, and as such, we provide detailed annotations on how to secure the correct numeral through specific operations. As no existing public datasets align with our task's unique characteristics, we annotated more than 27,000 numeral-rich news articles to further probe this research direction. These extensive annotations enable us to identify several unique characteristics of numerals in news headlines, thereby distinguishing our task settings from those of current numeral-related datasets.

We evaluate five models~\cite{Lewis2020,Raffel2020_T5,Zhang2019pegasus,wang2022season,liu2022brio} previously shown to perform well in headline generation tasks, conducting a human evaluation across three dimensions: numerical accuracy, reasonableness, and readability. Our findings suggest that alongside the traditional focuses of reasonableness and readability, there remains significant room for improvement in numerical accuracy. Through the release of our proposed NumHG dataset, we hope to accelerate progress in number-centric headline generation and stimulate further discussion on numeral-focused text generation.

\section{Related Work}
The task of headline generation, a form of text summarization, endeavors to condense a lengthy source text into a succinct summary. Text summarization approaches typically fall into two categories: extractive and abstractive. Extractive approaches involve selecting fitting sentences from the source text to serve as the summary, while abstractive approaches strive to create new sentences to encapsulate the source text. The concept of headline generation aligns more closely with abstractive methodologies. 

The emergence and development of large-scale pre-trained models~\cite{Raffel2020_T5,Lewis2020,Zhang2019pegasus} have notably advanced the capabilities of abstractive summarization models, to the extent that they now outperform extractive models. Some recent studies~\cite{dou2021gsum,wang2022season,liu2022brio} emphasize the significance of keyword sentences, asserting that these should be leveraged as guides for summary generation. GSum~\cite{dou2021gsum}, for example, initially performs extractive summarization, then incorporates the extractive summaries into the input for abstractive summarization. Despite experimental evidence supporting GSum's effectiveness, \citeauthor{wang2022season} \cite{wang2022season} argue that extractive summaries do not provide a reliable or flexible guide, potentially leading to information loss or noisy signals.

To tackle this issue, \textsc{Season}~\cite{wang2022season} adopts a dual approach, learning to predict the informativeness of each sentence and using this predicted information to guide abstractive summarization. Meanwhile, BRIO~\cite{liu2022brio} employs pre-trained abstractive models to generate candidate summaries, assigning each a probability mass according to their quality and defining a contrastive loss across the candidates. By considering both token-level prediction accuracy and sequence-level coordination, BRIO combines cross-entropy loss and contrastive loss for abstractive summarization.

Notably, the majority of these works focuses on the selection of words and the structure of sentences. However, our work diverges significantly as it specifically tackles the problem of numeral accuracy in headline generation—a factor often overlooked in other studies. Our newly proposed NumHG dataset, comprising over 27,000 annotated numeral-rich news articles, provides a valuable resource for enhancing the performance of numeral-aware headline generation tasks. The results from evaluating various models indicate a pressing need for improvement in numeral accuracy, which we believe will stimulate more focused research in this crucial yet neglected aspect of text generation.

\section{Dataset}
\subsection{Dataset Construction}
This section provides a comprehensive introduction to the proposed NumHG dataset.\footnote{\url{https://github.com/ArrowHuang/NumHG.git}} The primary source of our news articles is Newser\footnote{https://www.newser.com/}, a news aggregation platform that curates top stories from numerous U.S. and international outlets. Articles on Newser typically contain approximately 200 to 300 words.
Our focus for the NumHG dataset is news articles with numeral-infused headlines. Consequently, we eliminate articles without numerals in the headline. As a further restriction, NumHG is centered on headlines featuring only a single number, leading us to exclude articles with more than one numeral in the headline. These filtering processes result in a dataset of 27,746 instances.

\begin{table}
    \caption{
    Overview of predefined operators. $v$, $v_0$ and $v_1$ denote the selected numerals, and $e$ denotes the English word. $s$ and $c$ denote a span from the article and a constant, respectively.
    }
    \centering
    \resizebox{\linewidth}{!}{
    \begin{tabular}{lll}
    \hline
    \multicolumn{1}{c}{\textbf{Operator}} & \multicolumn{1}{c}{\textbf{Description}} & \multicolumn{1}{c}{\textbf{Ratio}}\\
    \hline
    Copy($v$) & Copy $v$ from the article & 65.00\% \\
    \hline
    Trans($e$) & Covert $e$ into a number & 17.37\% \\
    \hline
    \multirow{2}[0]{*}{Paraphrase($v_{0}$,$n$)}  & Paraphrase the form of digits to other & \multirow{2}[0]{*}{8.27\%} \\
    & representations &  \\
    \hline
    \multirow{2}[0]{*}{Round($v_{0}$,$c$)} & Hold $c$ digits after the decimal point & \multirow{2}[0]{*}{3.10\%} \\
     & of $v_{0}$ & \\
    \hline
    Subtract($v_{0}$,$v_{1}$) & Subtract $v_{1}$ from $v_{0}$ & 2.15\% \\
    \hline
    Add($v_{0}$,$v_{1}$)  & Add $v_{0}$ and $v_{1}$ & 1.73\% \\
    \hline
    Span($s$) & Select a span from the article & 1.34\% \\
    \hline
    Divide($v_{0}$,$v_{1}$) & Divide $v_{0}$ by $v_{1}$ & 0.54\% \\
    \hline
    Multiply($v_{0}$,$v_{1}$) & Multiply $v_{0}$ and $v_{1}$ & 0.50\% \\
    \hline
    \end{tabular}
    }
    \label{tab:operator-based representation}
\end{table}

\begin{table}[t]
\small
\caption{\label{Comparison of different MWP corpora}
Comparison of different corpora.}
\centering
    \begin{tabular}{lrrr} 
        \hline
        Corpus &  $\#$ Sents & $\#$ Words & $\#$ Nums \\ 
        \hline
        Dolphin18K~\cite{Huang2016} & 2.6 & 30.6 & 4.4   \\
        AQUA-RAT~\cite{Ling2017} & 2.2 & 32.5 & 4.2  \\
        Math23K~\cite{Wang2017} & 1.6 & 28.0 & 3.1  \\
        MathQA~\cite{Amini2019} & 2.0 & 37.9 & 4.5  \\
        SVAMP~\cite{patel2021nlp} & 2.8 & 31.8 & 3.2  \\
        \hline
        NumHG (Proposed)  & \textbf{9.4} & \textbf{191.8} & \textbf{13.7} \\
        \hline
    \end{tabular}
\end{table}

\begin{table*}[t]
    \caption{\label{tab:Headline Generation Results}
    Automatic evaluation results.} 
    \centering
    % \resizebox{0.8\linewidth}{!}{
    \begin{tabular}{llccc|ccc|ccc|c}
    \hline
          &  & \multicolumn{3}{c|}{Num Acc.} & \multicolumn{3}{c|}{ROUGE} & \multicolumn{3}{c|}{BERTScore} & \multicolumn{1}{c}{\multirow{2}[0]{*}{MoverScore}} \\
          \cline{3-11}
          &    &  \multicolumn{1}{c}{Overall} & \multicolumn{1}{c}{Copy} & \multicolumn{1}{c|}{Reasoning} & \multicolumn{1}{c}{1} & \multicolumn{1}{c}{2} & \multicolumn{1}{c|}{L}
          & \multicolumn{1}{c}{P} & \multicolumn{1}{c}{R} & \multicolumn{1}{c|}{F1}
          \\
    \hline
    BART &  & \textbf{70.09} & \textbf{73.88} & \textbf{61.54} & 46.63 & 21.79 & 41.55 & 48.02 & 49.19 & 48.62 & 62.57 \\ 
    T5 & & 67.84 & 71.42 & 59.74 & 47.82 & 23.10  & 42.89 & 50.23 & 49.64 & 49.94 & 62.98\\
    Pegasus &  & 66.45 & 70.25 & 57.86 & 48.08 & 23.40 & 43.25 & 50.97 & 49.99 & 50.49 & 63.11 \\ 
    \textsc{Season} &  & 67.81 & 71.11 & 60.35 & 48.58 & 23.81 & 43.74 & 51.64 & 50.32 & 50.98 & 63.29\\ 
    BRIO &  & 66.56 & 70.43 & 60.07 & \textbf{48.93} & \textbf{24.09} & \textbf{44.12} & \textbf{52.17} & \textbf{50.84} & \textbf{51.43} & \textbf{63.50}\\ 
    \hline
    \end{tabular}%
    % }
\end{table*}%d

\begin{table}[t]
    \caption{\label{tab:Human Evaluation Results}
    Human evaluation results.} 
    \centering
    \small
    \begin{tabular}{lccc}
    \hline
          & \multicolumn{1}{c}{Num Acc.} & \multicolumn{1}{c}{Reasonableness} & \multicolumn{1}{c}{Readability} \\
    \hline
    BART & 59.2 & 43.9 & 53.7 \\
    T5 & 53.9 & 52.1 & 55.9 \\
    Pegasus & 64.6 & 58.8 & 61.2 \\
    \textsc{Season} & 62.7 & 63.6 & 60.7 \\
    BRIO & \textbf{79.1}  & \textbf{65.2} & \textbf{63.5} \\
    \hline
    \end{tabular}%
\end{table}%

For accurate numeral generation in headlines, the model may need to manipulate the numerals in the article body or perform basic calculations. For instance, the headline numeral in the example provided in Table~\ref{tab:an example in NumHG} requires a simple calculation. Given the absence of suitable existing datasets for this purpose, we devise an annotation scheme to understand the operations between numerals in the news articles and the headlines. After sampling 3,000 instances for operator distribution analysis, we define a set of operators for our annotation guideline, as shown in Table~\ref{tab:operator-based representation}.
To derive the equations necessary for computing the correct numeral in the headline, we engage annotators via the \emph{Amazon Mechanical Turk} platform. We formulate a question by randomly omitting one number in the headline. The annotators are then presented with the news article and corresponding question, and they must determine whether the answer is inferable from the content. If the answer is unobtainable, annotators are required to provide a detailed reason, and we designate this instance as an unanswerable question. Conversely, if the answer can be inferred, annotators utilize the predefined operators, including \emph{Copy}, \emph{Trans}, \emph{Span}, \emph{Round}, \emph{Paraphrase}, \emph{Add}, \emph{Subtract}, \emph{Multiply}, and \emph{Divide}, to formulate an equation that yields the answer.

We enforce quality via an automated validation method. Given that the ground truth is formulated by professional journalists, we need to ensure that the annotator's equation aligns with it. When an annotator submits her/his equation, our program automatically calculates the result and checks for consistency with the article's numerals and text spans. In essence, an annotation will be successfully submitted if its result matches the ground truth and all used numbers and text spans appear in the article. Otherwise, annotators are prompted to review their work. While this automated method effectively filters obvious errors, it is incapable of distinguishing instances where all numbers are present in the article and the equation matches the ground truth. Thus, we deploy human validation to further verify the annotations in their context. For this task, we engage 840 experienced Turkers with a hit approval rate of no less than 85\% on the MTurk platform. We pay \$0.45 for each annotation, and each task is randomly assigned to three different annotators. An assignment is approved if at least two annotators concur on the answer. If a consensus is not reached, the assignment is reassigned to three new annotators.

\subsection{Dataset Analysis}
The proposed NumHG dataset is distinguished by three salient characteristics, as demonstrated in Table~\ref{Comparison of different MWP corpora}. Firstly, it exhibits considerably larger average sentence and word counts compared to its counterparts. Secondly, NumHG's source articles contain more numerals than those in preceding datasets. Finally, unlike other works, NumHG incorporates unanswerable questions, with annotators asked to provide a rationale for their unanswerability. This unique feature establishes a preliminary exploration of unanswerable questions in numeral problem-solving scenarios.
As depicted in Table~\ref{tab:operator-based representation}, the \emph{Copy} operator is the most commonly applied in the news articles within NumHG. The prevalence of simple operations (\emph{Copy}, \emph{Trans}, \emph{Span}, \emph{Round}, and \emph{Paraphrase}) underscores the journalistic practice of clear information delivery, avoiding any unnecessary challenge to the reader's numeracy skills. This also marks a notable departure from prior numerical reasoning datasets~\cite{Huang2016,Ling2017,Wang2017,Amini2019,patel2021nlp}, which predominantly aim to assess machine numeracy, thus not directly applicable to the news article context.

\begin{table*}[t]
\caption{ Case study on NumHG. }
\centering
\resizebox{\textwidth}{!}{
\begin{tabular}{|l|l|}
\hline
\multicolumn{2}{|p{\linewidth}|}{
(Aug 12, 2018 10:45 AM CDT) \textbf{Everyone in the town of Larrimah is under investigation for the disappearance of Paddy Moriarty—and that means all 11 people.} Authorities are poking around this dusty Australian pitstop after Moriarty, a day laborer, vanished along with his dog one night last December, the New York Times reports. \textbf{Top suspects in the presumed homicide include a pie-maker Moriarty hated, a gardener he argued with, and a bartender with a nasty tongue}.  He started abusing my customers, threatening tourists and scaring them away from [my] business,  said meat-pie cook Fran Hodgetts at an inquest last month, per ABC News Australia. Moriarty, 70, lived across from her Tea House eatery and got upset when her customers parked on his land. Like all players in this drama, 75-year-old Hodgetts denies guilt—but says she warned her burly gardener, Owen Laurie,  not to do anything stupid  after he argued with Moriarty about his barking dog three days before the disappearance. Laurie, 71, warned Moriarity to quiet the dog  or I'll shut it up for you,  per inquest testimony. Then there's former bartender Richard Simpson, one of the last residents to see Moriarty, who criticized the man but calls people who suspect Simpson  goddamn fools.  Jokes are circulating about Moriarty ending up in Hodgetts' meat pies, or fed to a hefty crocodile kept in town, but so far investigators appear stumped.  There have been a lot of problems in that community,  the detective in charge tells the Guardian.  But just because people argue doesn't mean they've gone out and killed him.
}
    \\ 
    \hline
     BART & All \textbf{11} People in This Town Are Top Suspects in Man's Disappearance \\
    \hline
     T5 & Everyone in Town Under Investigation for Disappearance of Man, \textbf{70} \\
    \hline
     Pegasus & \textbf{11} People Are Top Suspects for Man's Suspicious Death \\
    \hline
     \textsc{Season} & Everyone in This Town Is Under Investigation for Man's Disappearance \\
    \hline
     BRIO & \textbf{11} People Under Investigation in This Town for Missing Man \\
    \hline
    Ground Truth & Cops Probe Town of \textbf{11} People After Disappearance \\
    \hline
\end{tabular}
}
\label{tab:case study 1}
\end{table*}

\section{Experimental Evaluation}

\subsection{Experimental Setup}

\noindent
\textbf{Dataset Description}
To facilitate equitable comparisons, we employ 5-fold cross-validation on NumHG and report the averaged results. Each fold of the NumHG dataset is partitioned into 19,422 training pairs, 2,775 validation pairs, and 5,549 test pairs.

\noindent
\textbf{Models}
We evaluate a selection of robust baseline models on our proposed dataset. Specifically, we utilize \textit{BART}~\cite{Lewis2020}, \textit{T5}~\cite{Raffel2020_T5}, and \textit{Pegasus}~\cite{Zhang2019pegasus}, all renowned, large-scale, pre-trained sequence-to-sequence generation models. \textit{\textsc{Season}}~\cite{wang2022season} applies ROUGE-L between each document sentence and its corresponding reference summary to denote sentence informativeness, which subsequently guides abstractive summarization. \textit{BRIO}~\cite{liu2022brio} combines contrastive and cross-entropy losses to optimize both token-level prediction accuracy and sequence-level coordination.

\noindent
\textbf{Evaluation Metrics}
We employ ROUGE \cite{Rouge2004} as the automatic evaluation metric, incorporating ROUGE-1, ROUGE-2, and sentence-level ROUGE-L, using the rouge-score package\footnote{https://pypi.org/project/rouge-score/}. We also assess baseline performance using two model-based semantic similarity metrics, BERTScore~\cite{Zhang2020bertscore} and MoverScore~\cite{zhao2019moverscore}. Specifically, we use \textit{roberta-large} to calculate MoverScore and report the Precision (P), Recall (R), and F1 measure (F1) of BERTScore.

\noindent
\textbf{Implementation Details}
In our experiments, we fine-tune \textit{BART-large}, \textit{T5-large}, and \textit{Pegasus-large} from the \textit{transformers}~\cite{wolf2020transformers} library. We apply beam search with a beam size of 4 and set our batch size to 16 to fully utilize GPU memory. We employ the Adam optimizer~\cite{kingma2015adam} with a learning rate of 5e-5. The models are trained for 15 epochs in each fold, with the performance on the validation set guiding the checkpoint selection. For \textsc{Season}, we use \textit{BART-large} as a backbone, with all other settings consistent with the aforementioned. In the case of BRIO, we use \textit{BART-large} and employ beam search to generate 16 candidate summaries. The batch size is set to 16 for a total of 30 epochs, with the Adam optimizer and learning rate scheduling following the original paper's specifications. Additionally, we employ a linear warmup strategy, setting the number of warmup steps to 2,000. All experiments were conducted on 4 NVIDIA Tesla V100 (32G) GPUs.

\subsection{Experimental Results}
Our principal results are displayed in Table~\ref{tab:Headline Generation Results}. Among all the baseline models, BRIO exhibits superior performance on three summarization evaluation metrics: the ROUGE score, BERTScore, and MoverScore. \textsc{Season} utilizes the informativeness of each sentence to guide abstractive summarization, yielding promising improvements over the original BART by 1.95/2.02/2.19 points for ROUGE-1/2/L scores, 3.63/1.13/2.36 points for BERTScore-P/R/F1, and 0.72 points for MoverScore. Compared to Season, BRIO achieves marginal enhancements of 0.35/0.28/0.38 points for ROUGE-1/2/L scores, 0.53/0.52/0.45 points for BERTScore-P/R/F1, and 0.21 points for MoverScore.
We provide not only summarization evaluation scores but also numeral accuracy (Num Acc.) to assess whether the numerals generated in the headline match those in the ground truth. \emph{Overall} demonstrates performance on all test questions. \emph{Copy} denotes performance on questions where answers can be directly copied from the given article. \emph{Reasoning} pertains to questions that necessitate numerical reasoning to derive the answer. Although BRIO excels in three summarization evaluation metrics, Table~\ref{tab:Headline Generation Results} reveals that BART is the most effective in generating accurate numerals in the headline, with an overall numeral accuracy of 70.09$\%$.

\subsection{Human Evaluation}
We engaged five graduate students in communication and media studies as annotators, randomly selecting 100 instances from the NumHG test set. Each annotator was provided with the news article and five generated headlines, with no knowledge of which model generated which headline. We requested them to evaluate the generated headlines on three criteria:  \emph{Numeral Accuracy}, \emph{Reasonableness}, and \emph{Readability}. Numeral Accuracy assesses the correctness of numbers in the headline. The score is categorized as follows: 0 indicates all numbers in the generated headline are incorrect, 1 indicates a portion of numbers are correctly predicted, and 2 indicates all numbers are correctly predicted. Reasonableness requires the annotators to select the best headline for the given article. The best headline score 5, the second-best score 4, and the least favored score 1. Readability measures the ease or difficulty of understanding the headline. The readability score ranges between 1 and 5, where 1 signifies the generated headline is very challenging to read, and 5 indicates the headline is easily readable and understandable. Lastly, we represent the human evaluation results as percentages. For instance, we first sum up the numeral accuracy score of BART given by each evaluator. Then, we obtain 118.4, which is the average numeral accuracy score of the five annotators. Finally, we convert the numeral accuracy score into a percentage as 118.4/200, where 200 is the maximum possible score for the 100 sampled instances.

Table~\ref{tab:Human Evaluation Results} reports the human evaluation results. As illustrated in Table~\ref{tab:Human Evaluation Results}, BRIO excels in Numeral Accuracy, Reasonableness, and Readability in human evaluations. Interestingly, the numeral generated in the headline by baseline models can also be correct, even if it does not match the ground truth. However, a large number of numerals in generated headlines could be incorrect if the numeral's context is taken into account. In Table~\ref{tab:case study 1}, we present a case study. In this example, the numerals in the generated headlines by BART and Pegasus match the ground truth. Yet, when examining the news article, we find that 11 people in the town are under investigation with only 3 top suspects, including a pie-maker, a gardener, and a bartender. Therefore, the numerals in the headlines generated by BART and Pegasus are incorrect.

\section{Conclusion}
This paper concentrates on numerals when generating headlines and introduces a challenging dataset, NumHG. We employ several state-of-the-art models to generate headlines containing accurate numerals. However, experimental results indicate that these robust baseline models fail to generate accurate headlines with correct numerals. We will release NumHG under the CC BY-SA 4.0 license. In the future, we plan to inject numerical reasoning scheme to generation models to improve performance, and we also plan to design a better evaluation metric for number-focused text generation tasks.

\bibliographystyle{ACM-Reference-Format}
\bibliography{sample-base}


\begin{thebibliography}{17}

%%% the Web address.

\ifx \showCODEN    \undefined \def \showCODEN     #1{\unskip}     \fi
\ifx \showDOI      \undefined \def \showDOI       #1{#1}\fi
\ifx \showISBNx    \undefined \def \showISBNx     #1{\unskip}     \fi
\ifx \showISBNxiii \undefined \def \showISBNxiii  #1{\unskip}     \fi
\ifx \showISSN     \undefined \def \showISSN      #1{\unskip}     \fi
\ifx \showLCCN     \undefined \def \showLCCN      #1{\unskip}     \fi
\ifx \shownote     \undefined \def \shownote      #1{#1}          \fi
\ifx \showarticletitle \undefined \def \showarticletitle #1{#1}   \fi
\ifx \showURL      \undefined \def \showURL       {\relax}        \fi
% invisible to TeX
\providecommand\bibfield[2]{#2}
\providecommand\bibinfo[2]{#2}
\providecommand\natexlab[1]{#1}
\providecommand\showeprint[2][]{arXiv:#2}

\bibitem[\protect\citeauthoryear{Amini, Gabriel, Lin, Koncel-Kedziorski, Choi,
  and Hajishirzi}{Amini et~al\mbox{.}}{2019}]%
        {Amini2019}
\bibfield{author}{\bibinfo{person}{Aida Amini}, \bibinfo{person}{Saadia
  Gabriel}, \bibinfo{person}{Peter Lin}, \bibinfo{person}{Rik
  Koncel-Kedziorski}, \bibinfo{person}{Yejin Choi}, {and}
  \bibinfo{person}{Hannaneh Hajishirzi}.} \bibinfo{year}{2019}\natexlab{}.
\newblock \showarticletitle{MathQA: Towards Interpretable Math Word Problem
  Solving with Operation-Based Formalisms}. In
  \bibinfo{booktitle}{\emph{Proceedings of the 2019 Conference of the North
  American Chapter of the Association for Computational Linguistics}}.
  \bibinfo{pages}{2357--2367}.
\newblock


\bibitem[\protect\citeauthoryear{Dou, Liu, Hayashi, Jiang, and Neubig}{Dou
  et~al\mbox{.}}{2021}]%
        {dou2021gsum}
\bibfield{author}{\bibinfo{person}{Zi-Yi Dou}, \bibinfo{person}{Pengfei Liu},
  \bibinfo{person}{Hiroaki Hayashi}, \bibinfo{person}{Zhengbao Jiang}, {and}
  \bibinfo{person}{Graham Neubig}.} \bibinfo{year}{2021}\natexlab{}.
\newblock \showarticletitle{GSum: A General Framework for Guided Neural
  Abstractive Summarization}. In \bibinfo{booktitle}{\emph{Proceedings of the
  2021 Conference of the North American Chapter of the Association for
  Computational Linguistics: Human Language Technologies}}.
  \bibinfo{pages}{4830--4842}.
\newblock


\bibitem[\protect\citeauthoryear{Huang, Shi, Lin, Yin, and Ma}{Huang
  et~al\mbox{.}}{2016}]%
        {Huang2016}
\bibfield{author}{\bibinfo{person}{Danqing Huang}, \bibinfo{person}{Shuming
  Shi}, \bibinfo{person}{Chin-Yew Lin}, \bibinfo{person}{Jian Yin}, {and}
  \bibinfo{person}{Wei-Ying Ma}.} \bibinfo{year}{2016}\natexlab{}.
\newblock \showarticletitle{How well do Computers Solve Math Word Problems?
  Large-Scale Dataset Construction and Evaluation}. In
  \bibinfo{booktitle}{\emph{Proceedings of the 54th Annual Meeting of the
  Association for Computational Linguistics}}. \bibinfo{pages}{887--896}.
\newblock


\bibitem[\protect\citeauthoryear{Kingma and Ba}{Kingma and Ba}{2015}]%
        {kingma2015adam}
\bibfield{author}{\bibinfo{person}{Diederik~P. Kingma} {and}
  \bibinfo{person}{Jimmy Ba}.} \bibinfo{year}{2015}\natexlab{}.
\newblock \showarticletitle{Adam: {A} Method for Stochastic Optimization}. In
  \bibinfo{booktitle}{\emph{3rd International Conference on Learning
  Representations, {ICLR} 2015, San Diego, CA, USA, May 7-9, 2015, Conference
  Track Proceedings}}, \bibfield{editor}{\bibinfo{person}{Yoshua Bengio} {and}
  \bibinfo{person}{Yann LeCun}} (Eds.).
\newblock


\bibitem[\protect\citeauthoryear{Lewis, Liu, Goyal, Ghazvininejad, Mohamed,
  Levy, Stoyanov, and Zettlemoyer}{Lewis et~al\mbox{.}}{2020}]%
        {Lewis2020}
\bibfield{author}{\bibinfo{person}{Mike Lewis}, \bibinfo{person}{Yinhan Liu},
  \bibinfo{person}{Naman Goyal}, \bibinfo{person}{Marjan Ghazvininejad},
  \bibinfo{person}{Abdelrahman Mohamed}, \bibinfo{person}{Omer Levy},
  \bibinfo{person}{Veselin Stoyanov}, {and} \bibinfo{person}{Luke
  Zettlemoyer}.} \bibinfo{year}{2020}\natexlab{}.
\newblock \showarticletitle{{BART}: Denoising Sequence-to-Sequence Pre-training
  for Natural Language Generation, Translation, and Comprehension}. In
  \bibinfo{booktitle}{\emph{Proceedings of the 58th Annual Meeting of the
  Association for Computational Linguistics}}. \bibinfo{pages}{7871--7880}.
\newblock


\bibitem[\protect\citeauthoryear{Lin}{Lin}{2004}]%
        {Rouge2004}
\bibfield{author}{\bibinfo{person}{Chin-Yew Lin}.}
  \bibinfo{year}{2004}\natexlab{}.
\newblock \showarticletitle{{ROUGE}: A Package for Automatic Evaluation of
  Summaries}. In \bibinfo{booktitle}{\emph{Text Summarization Branches Out}}.
  \bibinfo{publisher}{Association for Computational Linguistics},
  \bibinfo{address}{Barcelona, Spain}, \bibinfo{pages}{74--81}.
\newblock


\bibitem[\protect\citeauthoryear{Ling, Yogatama, Dyer, and Blunsom}{Ling
  et~al\mbox{.}}{2017}]%
        {Ling2017}
\bibfield{author}{\bibinfo{person}{Wang Ling}, \bibinfo{person}{Dani Yogatama},
  \bibinfo{person}{Chris Dyer}, {and} \bibinfo{person}{Phil Blunsom}.}
  \bibinfo{year}{2017}\natexlab{}.
\newblock \showarticletitle{Program Induction by Rationale Generation: Learning
  to Solve and Explain Algebraic Word Problems}. In
  \bibinfo{booktitle}{\emph{Proceedings of the 55th Annual Meeting of the
  Association for Computational Linguistics}}. \bibinfo{pages}{158--167}.
\newblock


\bibitem[\protect\citeauthoryear{Liu, Liu, Radev, and Neubig}{Liu
  et~al\mbox{.}}{2022}]%
        {liu2022brio}
\bibfield{author}{\bibinfo{person}{Yixin Liu}, \bibinfo{person}{Pengfei Liu},
  \bibinfo{person}{Dragomir Radev}, {and} \bibinfo{person}{Graham Neubig}.}
  \bibinfo{year}{2022}\natexlab{}.
\newblock \showarticletitle{{BRIO}: Bringing Order to Abstractive
  Summarization}. In \bibinfo{booktitle}{\emph{Proceedings of the 60th Annual
  Meeting of the Association for Computational Linguistics (Volume 1: Long
  Papers)}}. \bibinfo{publisher}{Association for Computational Linguistics},
  \bibinfo{address}{Dublin, Ireland}, \bibinfo{pages}{2890--2903}.
\newblock


\bibitem[\protect\citeauthoryear{Matsumaru, Takase, and Okazaki}{Matsumaru
  et~al\mbox{.}}{2020}]%
        {matsumaru-etal-2020-improving}
\bibfield{author}{\bibinfo{person}{Kazuki Matsumaru}, \bibinfo{person}{Sho
  Takase}, {and} \bibinfo{person}{Naoaki Okazaki}.}
  \bibinfo{year}{2020}\natexlab{}.
\newblock \showarticletitle{Improving Truthfulness of Headline Generation}. In
  \bibinfo{booktitle}{\emph{Proceedings of the 58th Annual Meeting of the
  Association for Computational Linguistics}}. \bibinfo{publisher}{Association
  for Computational Linguistics}, \bibinfo{address}{Online},
  \bibinfo{pages}{1335--1346}.
\newblock
\urldef\tempurl%
\url{https://doi.org/10.18653/v1/2020.acl-main.123}
\showDOI{\tempurl}


\bibitem[\protect\citeauthoryear{Patel, Bhattamishra, and Goyal}{Patel
  et~al\mbox{.}}{2021}]%
        {patel2021nlp}
\bibfield{author}{\bibinfo{person}{Arkil Patel}, \bibinfo{person}{Satwik
  Bhattamishra}, {and} \bibinfo{person}{Navin Goyal}.}
  \bibinfo{year}{2021}\natexlab{}.
\newblock \showarticletitle{Are NLP Models really able to Solve Simple Math
  Word Problems?}. In \bibinfo{booktitle}{\emph{Proceedings of the 2021
  Conference of the North {A}merican Chapter of the Association for
  Computational Linguistics: Human Language Technologies}}.
  \bibinfo{pages}{2080--2094}.
\newblock


\bibitem[\protect\citeauthoryear{Raffel, Shazeer, Roberts, Lee, Narang, Matena,
  Zhou, Li, and Liu}{Raffel et~al\mbox{.}}{2020}]%
        {Raffel2020_T5}
\bibfield{author}{\bibinfo{person}{Colin Raffel}, \bibinfo{person}{Noam
  Shazeer}, \bibinfo{person}{Adam Roberts}, \bibinfo{person}{Katherine Lee},
  \bibinfo{person}{Sharan Narang}, \bibinfo{person}{Michael Matena},
  \bibinfo{person}{Yanqi Zhou}, \bibinfo{person}{Wei Li}, {and}
  \bibinfo{person}{Peter~J. Liu}.} \bibinfo{year}{2020}\natexlab{}.
\newblock \showarticletitle{Exploring the Limits of Transfer Learning with a
  Unified Text-to-Text Transformer}.
\newblock \bibinfo{journal}{\emph{J. Mach. Learn. Res.}}  \bibinfo{volume}{21}
  (\bibinfo{year}{2020}), \bibinfo{pages}{140:1--140:67}.
\newblock


\bibitem[\protect\citeauthoryear{Wang, Song, Zhang, Jin, Cho, Yao, Wang, Chen,
  and Yu}{Wang et~al\mbox{.}}{2022}]%
        {wang2022season}
\bibfield{author}{\bibinfo{person}{Fei Wang}, \bibinfo{person}{Kaiqiang Song},
  \bibinfo{person}{Hongming Zhang}, \bibinfo{person}{Lifeng Jin},
  \bibinfo{person}{Sangwoo Cho}, \bibinfo{person}{Wenlin Yao},
  \bibinfo{person}{Xiaoyang Wang}, \bibinfo{person}{Muhao Chen}, {and}
  \bibinfo{person}{Dong Yu}.} \bibinfo{year}{2022}\natexlab{}.
\newblock \showarticletitle{Salience Allocation as Guidance for Abstractive
  Summarization}. In \bibinfo{booktitle}{\emph{Proceedings of the 2022
  Conference on Empirical Methods in Natural Language Processing}}.
  \bibinfo{publisher}{Association for Computational Linguistics},
  \bibinfo{address}{Abu Dhabi, United Arab Emirates},
  \bibinfo{pages}{6094--6106}.
\newblock


\bibitem[\protect\citeauthoryear{Wang, Liu, and Shi}{Wang
  et~al\mbox{.}}{2017}]%
        {Wang2017}
\bibfield{author}{\bibinfo{person}{Yan Wang}, \bibinfo{person}{Xiaojiang Liu},
  {and} \bibinfo{person}{Shuming Shi}.} \bibinfo{year}{2017}\natexlab{}.
\newblock \showarticletitle{Deep Neural Solver for Math Word Problems}. In
  \bibinfo{booktitle}{\emph{Proceedings of the 2017 Conference on Empirical
  Methods in Natural Language Processing}}. \bibinfo{pages}{845--854}.
\newblock


\bibitem[\protect\citeauthoryear{Wolf, Debut, Sanh, Chaumond, Delangue, Moi,
  Cistac, Rault, Louf, Funtowicz, Davison, Shleifer, von Platen, Ma, Jernite,
  Plu, Xu, Scao, Gugger, Drame, Lhoest, and Rush}{Wolf et~al\mbox{.}}{2020}]%
        {wolf2020transformers}
\bibfield{author}{\bibinfo{person}{Thomas Wolf}, \bibinfo{person}{Lysandre
  Debut}, \bibinfo{person}{Victor Sanh}, \bibinfo{person}{Julien Chaumond},
  \bibinfo{person}{Clement Delangue}, \bibinfo{person}{Anthony Moi},
  \bibinfo{person}{Pierric Cistac}, \bibinfo{person}{Tim Rault},
  \bibinfo{person}{R{\'{e}}mi Louf}, \bibinfo{person}{Morgan Funtowicz},
  \bibinfo{person}{Joe Davison}, \bibinfo{person}{Sam Shleifer},
  \bibinfo{person}{Patrick von Platen}, \bibinfo{person}{Clara Ma},
  \bibinfo{person}{Yacine Jernite}, \bibinfo{person}{Julien Plu},
  \bibinfo{person}{Canwen Xu}, \bibinfo{person}{Teven~Le Scao},
  \bibinfo{person}{Sylvain Gugger}, \bibinfo{person}{Mariama Drame},
  \bibinfo{person}{Quentin Lhoest}, {and} \bibinfo{person}{Alexander~M. Rush}.}
  \bibinfo{year}{2020}\natexlab{}.
\newblock \showarticletitle{Transformers: State-of-the-Art Natural Language
  Processing}. In \bibinfo{booktitle}{\emph{Proceedings of the 2020 Conference
  on Empirical Methods in Natural Language Processing: System Demonstrations,
  {EMNLP} 2020 - Demos, Online, November 16-20, 2020}}.
  \bibinfo{publisher}{Association for Computational Linguistics},
  \bibinfo{pages}{38--45}.
\newblock


\bibitem[\protect\citeauthoryear{Zhang, Zhao, Saleh, and Liu}{Zhang
  et~al\mbox{.}}{2020b}]%
        {Zhang2019pegasus}
\bibfield{author}{\bibinfo{person}{Jingqing Zhang}, \bibinfo{person}{Yao Zhao},
  \bibinfo{person}{Mohammad Saleh}, {and} \bibinfo{person}{Peter Liu}.}
  \bibinfo{year}{2020}\natexlab{b}.
\newblock \showarticletitle{{PEGASUS}: Pre-training with Extracted
  Gap-sentences for Abstractive Summarization}. In
  \bibinfo{booktitle}{\emph{Proceedings of the 37th International Conference on
  Machine Learning}}. \bibinfo{pages}{11328--11339}.
\newblock


\bibitem[\protect\citeauthoryear{Zhang, Kishore, Wu, Weinberger, and
  Artzi}{Zhang et~al\mbox{.}}{2020a}]%
        {Zhang2020bertscore}
\bibfield{author}{\bibinfo{person}{Tianyi Zhang}, \bibinfo{person}{Varsha
  Kishore}, \bibinfo{person}{Felix Wu}, \bibinfo{person}{Kilian~Q. Weinberger},
  {and} \bibinfo{person}{Yoav Artzi}.} \bibinfo{year}{2020}\natexlab{a}.
\newblock \showarticletitle{BERTScore: Evaluating Text Generation with BERT}.
  In \bibinfo{booktitle}{\emph{International Conference on Learning
  Representations}}.
\newblock


\bibitem[\protect\citeauthoryear{Zhao, Peyrard, Liu, Gao, Meyer, and Eger}{Zhao
  et~al\mbox{.}}{2019}]%
        {zhao2019moverscore}
\bibfield{author}{\bibinfo{person}{Wei Zhao}, \bibinfo{person}{Maxime Peyrard},
  \bibinfo{person}{Fei Liu}, \bibinfo{person}{Yang Gao},
  \bibinfo{person}{Christian~M. Meyer}, {and} \bibinfo{person}{Steffen Eger}.}
  \bibinfo{year}{2019}\natexlab{}.
\newblock \showarticletitle{MoverScore: Text Generation Evaluating with
  Contextualized Embeddings and Earth Mover Distance}. In
  \bibinfo{booktitle}{\emph{Proceedings of the 2019 Conference on Empirical
  Methods in Natural Language Processing}}. \bibinfo{publisher}{Association for
  Computational Linguistics}, \bibinfo{address}{Hong Kong, China}.
\newblock


\end{thebibliography}

\end{document}